%% file: main.tex
\documentclass[conference]{IEEEtran}
\IEEEoverridecommandlockouts
\usepackage{cite}
\usepackage{amsmath,amssymb,amsfonts}
\usepackage{algorithmic}
\usepackage{graphicx}

\usepackage{textcomp}
\usepackage{booktabs}
\usepackage{accessibility}
\usepackage{multirow}
\usepackage{bm}
\usepackage[table]{xcolor}
\usepackage{hyperref}
\usepackage{algorithm}
\usepackage{algorithmic}
\usepackage{pifont} 
\def\BibTeX{{\rm B\kern-.05em{\sc i\kern-.025em b}\kern-.08em
    T\kern-.1667em\lower.7ex\hbox{E}\kern-.125emX}}
\begin{document}

\title{\textsc{ChainsFormer}: Numerical Reasoning on Knowledge Graphs from a Chain Perspective}


\author{
    \IEEEauthorblockN{
        Ze Zhao\textsuperscript{1}, 
        Bin Lu\textsuperscript{1}, 
        Xiaoying Gan\textsuperscript{1,*\footnotemark},
        Gu Tang\textsuperscript{1}, 
        Luoyi Fu\textsuperscript{2}, 
        Xinbing Wang\textsuperscript{1},
    }
    \IEEEauthorblockA{
        \textsuperscript{1} \textit{Department of Electronic Engineering, Shanghai Jiao Tong University, Shanghai, China} \\
        \textsuperscript{2} \textit{Department of Computer Science, Shanghai Jiao Tong University, Shanghai, China} \\
        }
    \IEEEauthorblockA{
        \textsuperscript{1}
        \texttt{\{zhaoze, robinlu1209, ganxiaoying, gutang, xwang8\}@sjtu.edu.cn}, \\
        \textsuperscript{2}
        \texttt{yiluofu@sjtu.edu.cn},
        \hspace{1em}
    }
}

\maketitle

\begin{abstract}
Reasoning over Knowledge Graphs (KGs) plays a pivotal role in knowledge graph completion or question answering systems, providing richer and more accurate triples and attributes. As numerical attributes become increasingly essential in characterizing entities and relations in KGs, the ability to reason over these attributes has gained significant importance. Existing graph-based methods such as Graph Neural Networks (GNNs) and Knowledge Graph Embeddings (KGEs), primarily focus on aggregating homogeneous local neighbors and implicitly embedding diverse triples. However, these approaches often fail to fully leverage the potential of logical paths within the graph, limiting their effectiveness in exploiting the reasoning process. To address these limitations, we propose ChainsFormer, a novel chain-based framework designed to support numerical reasoning. Chainsformer not only explicitly constructs logical chains but also expands the reasoning depth to multiple hops. Specially, we introduces Relation-Attribute Chains (RA-Chains), a specialized logic chain, to model sequential reasoning patterns. ChainsFormer captures the step-by-step nature of multi-hop reasoning along RA-Chains by employing sequential in-context learning. To mitigate the impact of noisy chains, we propose a hyperbolic affinity scoring mechanism that selects relevant logic chains in a variable-resolution space. Furthermore, ChainsFormer incorporates an attention-based numerical reasoner to identify critical reasoning paths, enhancing both reasoning accuracy and transparency. Experimental results demonstrate that ChainsFormer significantly outperforms state-of-the-art methods, achieving up to a 20.0\% improvement in performance. The implementations are available at \href{https://github.com/zhaodazhuang2333/ChainsFormer}{https://github.com/zhaodazhuang2333/ChainsFormer.}

\renewcommand{\thefootnote}{\fnsymbol{footnote}} 
\footnotetext[1]{Corresponding author: Xiaoying Gan.}

\end{abstract}

\begin{IEEEkeywords}
Knowledge graph, Numerical reasoning, Chain of thought.
\end{IEEEkeywords}

\input{1_introduction}

\input{2_related_work}

\input{3_preliminaries}

\input{4_methodology}

\input{5_experiment}

\input{6_conclusion}
\input{7_Acknowledgement}

\bibliographystyle{unsrt}
\bibliography{main}

\vspace{12pt}

\end{document}

%% file: 1_introduction.tex
\section{INTRODUCTION}
\label{1_introduction}

Reasoning on Knowledge Graphs (KGs) facilitates the inference of unknown information and knowledge, significantly enhancing the utility and applicability of KGs. Recently, it has been widely adopted in various applications, including knowledge graph completion~\cite{rotate,ULTRA}, recommendation~\cite{KGrec,editkg}, and question answering~\cite{kbqa1,kbqa2}, demonstrating impressive achievements. These methods utilize triples in KGs to uncover hidden patterns, predict missing links, and provide context-aware recommendations~\cite{Acemap}. Furthermore, reasoning on KGs can assist LLMs improve domain-specific expertise and mitigate hallucination issues~\cite{kgsurvey_llm, ToG}. As a result, reasoning over relational triples has gained considerable attention.

\begin{figure}
    \centering
    \includegraphics[width=0.95\linewidth]{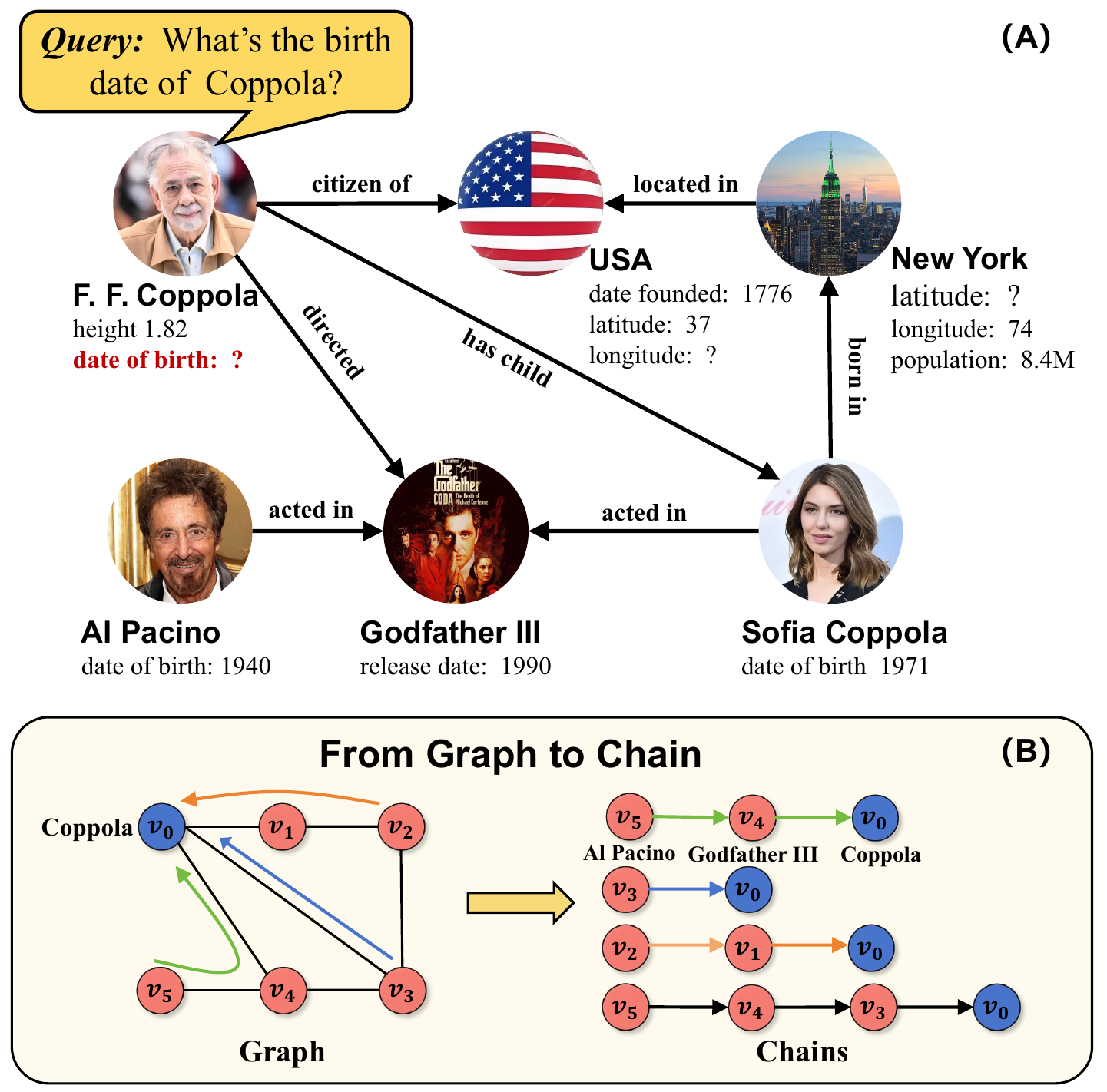}
    \caption{An example of numerical reasoning: predicting missing numerical attributes (e.g., birth date) using  relations and known attributes in a KG. Our exploration shifts from a graph-based view to a chain-based perspective for numerical reasoning.}
    \label{fig:intro}
    \vspace{-10pt}
\end{figure}

Whereas, reasoning over numerical attributes, i.e. numerical reasoning, remains an under explored area. Numerical reasoning focuses on predicting unknown numerical attributes based on existing information in KGs. For instance, as depicted in Fig~\ref{fig:intro}(A), when querying the missing attribute, Coppola's birth date, the answer cannot be directly retrieved from neighboring entities. Instead, it necessitates aggregating attributes from relevant entities, including local neighbors like Godfather III, Sofia Coppola, as well as multi-hop neighbors like Al Pacino. Hereby, such numerical reasoning process is critical for tasks like knowledge graph completion and question answering~\cite{PLM}, providing a more comprehensive characterization of entity properties~\cite{RAKGE}. This, in turn, enriches the domain knowledge available to large language models (LLMs), helping to mitigate hallucination issues and enhancing their overall reliability.

Existing methods for numerical reasoning on KGs primarily rely on entities and relations in a graph-based manner, which can be broadly categorized into GNN-based and KGE-based approaches. 
GNN-based methods~\cite{MTKGNN},\cite{MrAP} utilize graph neural networks to aggregate information from neighboring nodes and relations, effectively capturing local structural patterns and relational dependencies. KGE-based methods~\cite{KGA},\cite{HyNT} embed entities and relations into a shared latent space, enabling the discovery of implicit connections between attributes. However, these methods tend to aggregate information from neighbors in a homogeneous manner, failing to capture the full potential of logical paths and distinguish relevant reasoning paths from numerous noisy paths, affecting their performance in complex numerical reasoning tasks.

Recently, to explore the benefits of reasoning in complex tasks, Chain of Thought (CoT) has been proposed as a promising paradigm in a multi-hop scenario. It is initially designed to address challenges in natural language processing, such as numerical problem-solving~\cite{CoT1} and logic-based question answering~\cite{CoT2}. By breaking down problems into sequential reasoning steps, CoT enables the construction of logic chains that explicitly capture intermediate processes~\cite{CoT3}. This approach significantly improves performance in tasks requiring logical consistency and numerical accuracy~\cite{CoT4},\cite{CoT5}.

\begin{figure}
    \centering
    \includegraphics[width=0.9\linewidth]{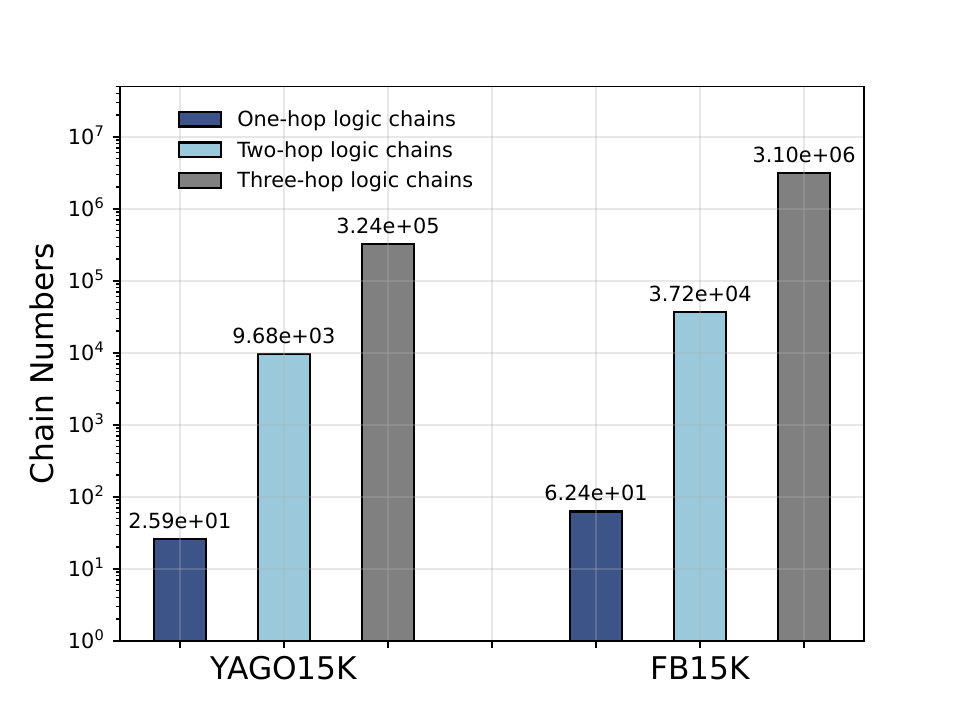}
    \caption{Average number of logic chains connected to each query in YAGO15K and FB15K datasets. The number of logic chains increases significantly with hops, reaching 3.24e+5 for YAGO15K and 3.10e+6 for FB15K at three hops, underscoring the challenges for large-scale, long-distance reasoning.}
    \label{fig:intro_fig_2}
    \vspace{-15pt}
\end{figure}

Motivated by the principles of Chain of Thought, we propose a novel paradigm based on chain-based reasoning rather than graph-based reasoning. Instead of reasoning homogeneously over static graph structures, we focus on selecting relevant logic chains to precisely profile and explicitly capture reasoning patterns. For instance, as illustrated in Fig.~\ref{fig:intro}(B), one such chain connects Al Pacino, Godfather III, and Coppola, providing a multi-hop reasoning path for the query. Furthermore, we represent these chains sequentially to align with the step-by-step nature of multi-hop reasoning, enabling better reasoning capability. However, this transition introduces two key challenges:

(1) \textit{How to design logic chains to effectively capture reasoning patterns?}
Designing logic chains requires capturing the sequential dependencies and reasoning patterns inherent in numerical tasks. This involves defining structures that can be tokenized into reasoning-relevant components, while preserving their logical order and interdependencies. Appropriate profiling and representation of these chains is crucial to reflect the step-by-step nature of reasoning, ensuring the model can handle diverse chains.

(2)\textit{ How to filter out irrelevant logic chains?}
The number of possible logic chains grows exponentially with the depth of reasoning, as shown in Figure~\ref{fig:intro_fig_2}. This explosion leads to many irrelevant or noisy logic chains. Identifying and retaining the relevant ones is crucial for numerical reasoning. Accurately selecting relevant logic chains from the vast search space based on the reasoning task remains highly challenging.

To tackle these challenges, we propose \textbf{ChainsFormer}, a novel framework for numerical reasoning from a chain perspective. 
ChainsFormer constructs a Tree of Chains (ToC) based on query-guided retrieval over the graph.
For the first challenge, ChainsFormer introduces Relation-Attribute Chain (RA-Chain), which profiles logic chains as sequences of composable units (attributes and relations). RA-Chains enable an in-context representation that captures step-by-step reasoning processes and integrates numerical features, allowing sequence embedding techniques to learn diverse reasoning patterns.
For the second challenge, ChainsFormer incorporates a hyperbolic affinity scoring mechanism to assess the relevance of RA-Chains. Leveraging the hierarchical structure of ToCs, this mechanism effectively filters out irrelevant or noisy paths in variable resolution space, ensuring accurate reasoning. Additionally, an attention-based numerical reasoner evaluates the contributions of individual chains, enhancing transparency by identifying key reasoning paths that lead to predictions.

In summary, the main contributions of our work are as follows:
\begin{itemize}

\item We propose ChainsFormer, a novel framework for numerical reasoning on knowledge graphs. By introducing Relation Attribute Chains (RA-Chains), ChainsFormer can precisely profile and capture sequential reasoning patterns. It  extends the depth of reasoning to multiple hops.

\item  We construct a Tree of Chains (ToC) via query-guided retrieval to manage RA-Chains and utilize an in-context representation to enable the step-by-step reasoning process along the chains. This representation enhances the logical consistency and flexibility of reasoning over long and multi-hops.

\item We design a hyperbolic affinity scoring mechanism that leverages the hierarchical structure of ToCs to effectively filter out irrelevant chains in variable resolution space. This mechanism reduces irrelevant and homogeneous propagation, improving the transparency and accuracy of numerical reasoning.


\end{itemize}

Experimental results demonstrate that our proposed approach achieves the best performance compared to existing methods, with improvements ranging from 7.4\% to 20.0\% over the state-of-the-art.

%% file: 2_related_work.tex
\section{RELATED WORK}
\label{2_related_work}

\subsection{Reasoning on Knowledge Graphs} Knowledge Graph Reasoning (KGR) aims to infer new knowledge from existing data in KGs~\cite{knowledgereasoningreview}. It is widely used in applications like question answering~\cite{kbqa1, kbqa2, kgqa3}, recommendation systems~\cite{KGrec,editkg}, and knowledge graph completion~\cite{transe}. KGR techniques are generally categorized into rule-based reasoning~\cite{rule_model2,rule4, rule_kgr}, Knowledge Graph Embedding (KGE) methods. Among these methods, KGE methods, such as the Trans series~\cite{transe,TransH,TransR,rotate} and tensor decomposition models~\cite{DistMult,ComplEx}, have gained significant attention due to their expressive power. The rise of Graph Neural Networks (GNNs)~\cite{GCN,hgt} leads to a focus on GNN-based KGR~\cite{CompGCN, NBFNet, ULTRA}, showcasing broader applications. While most existing KGR methods focus on relational triples, some recent studies explore multimodal information in KGs~\cite{multi_modal_1, multi_modal_2}. However, reasoning over numerical information remains largely overlooked. Only some KGE methods incorporate numerical attributes to enhance KG completion. For example, TransEA~\cite{TransEA} and Marine~\cite{Marine} modify loss functions, while LiteralE~\cite{LiteralE} and LiteralC~\cite{LiteralC} use gating mechanisms to integrate entities and attributes. Other approaches, like NRN~\cite{NRN}, encode numerical distributions for unified reasoning, and RAKGE~\cite{RAKGE} employs relation-aware encoders for numerical information. Despite these advances, existing methods focus on reasoning over predefined values and cannot infer or complete missing numerical attributes in KGs.

\subsection{Numerical Reasoning on Knowledge Graphs} Numerical reasoning on KGs is first introduced by MTKGNN~\cite{MTKGNN}, and then evolves into two main categories: KGE based methods and graph neural networks (GNN) based methods. KGE based method uses embedding directly for numerical reasoning. For instance, ~\cite{PLM} integrates textual features from pre-train language models for regression on numerical attributes, while HyNT~\cite{HyNT} treats numerical attributes as qualifiers of triples, combining triple information for numerical prediction. However, directly regressing sparse numerical attributes presents challenges~\cite{nlp_survey}. Although KGA~\cite{KGA} simplifies this task to link prediction by using binning, the inherent quantization error in binning necessitates a trade-off between classification difficulty and quantization precision. GNN-based methods show more promise, yet current approaches remain rudimentary. NAP++~\cite{NAP++} aggregates attributes directly via nearest K-NN, whereas MrAP~\cite{MrAP} considers attribute propagation under different paths but is confined to local neighbors. Our method, ChainsFormer, shifts focus from graph-based to chain-based perspective, achieving numerical reasoning in knowledge graphs based on multi-hop path reasoning.

\subsection{Hyperbolic Embedding} Compared to Euclidean space, hyperbolic geometry grows exponentially with its radius, making it a promising alternative for handling graph data characterized by tree-like structures or power-law distributions~\cite{hypersurvey, hypergnn}. Hyperbolic Neural Networks ~\cite{ganea2018hyperbolic} pioneered the adaptation of essential deep learning tools to hyperbolic space. ~\cite{hgnn} first extended graph neural networks into hyperbolic space using tangent space, catalyzing subsequent research in hyperbolic graph representation~\cite{hgnn21, hgnn23}. MuRP~\cite{MuRP}, ATTH~\cite{ATTH}, and GIE~\cite{GIE} which extended hyperbolic geometry to knowledge graphs, effectively capturing latent hierarchical structures and modeling hierarchical relationships between facts. However, existing studies have not yet addressed the embedding of logic chains composed of multi-hop relations and attributes in hyperbolic space.

%% file: 3_preliminaries.tex
\section{PRELIMINARY}
\label{3_preliminaries}
In this section, we present the problem formulation to numerical reasoning and a brief introduction of hyperbolic geometry, which will be used in ChainsFormer.

\subsection{Problem Formulation}

To precisely describe the problem, we mathematically define numerical reasoning problems on knowledge graphs. A multi-relational KG enriched with numerical attributes is defined as $\mathcal{G} = (\mathcal{V}, \mathcal{R}, \mathcal{A}, \mathcal{N}) $, where $\mathcal{V}$ denotes the set of nodes (entities), $\mathcal{R}$ is the set of relation types, $\mathcal{A}$ is the set of numerical attribute types, and $\mathcal{N}$ is the set of numerical values. 
In parallel with relational triples $\mathcal{E}_r \subset (\mathcal{V} \times \mathcal{R} \times \mathcal{V})$, we define numerical facts in the KG as numerical triples $\mathcal{E}_a \subset (\mathcal{V} \times \mathcal{A} \times \mathcal{N})$. 

\textbf{Definition 1. \textit{Numerical Reasoning on Knowledge Graphs.}} \textit{Given a multi-relational knowledge graph $\mathcal{G} = (\mathcal{V}, \mathcal{R}, \mathcal{A}, \mathcal{N})$, numerical reasoning aims to infer missing numerical values within the knowledge graph $\mathcal{G}$. Let $v \in \mathcal{V}$ denote the query entity,  $a \in \mathcal{A}$ represent the numerical attribute, and $n \in \mathcal{N}$ signify the predicted numerical value. The numerical reasoning task is formulated as a regression problem, where the goal is to learn a function $f: \mathcal{G} \times \mathcal{V} \times \mathcal{A} \to \mathcal{N}$ that predicts the numerical value $n$ for the incomplete triple  $(v, a, ?)$.}

\subsection{Hyperbolic Geometry}

Hyperbolic geometry is a non-Euclidean geometry characterized by constant negative curvature~\cite{nickel2017poincare}. In hyperbolic space, the exponential growth of distances aligns with the exponential growth of nodes in a tree structure~\cite{ganea2018hyperbolic}, making it ideal for modeling hierarchical and branching structures. We select the $d$-dimensional Poincar\'{e} ball with negative curvature $-\mathfrak{c}(\mathfrak{c}>0):\mathbb{B}^{d,\mathfrak{c}} = \{\bm{x}\in \mathbb{R}^d:\Vert \bm{x}\Vert ^2<\frac{1}{\mathfrak{c}}\}$. Many operations well-defined in Euclidean space are not applicable in hyperbolic space. Therefore, we introduce the following key concepts for hyperbolic geometry: M$\ddot{o}$bius addition and hyperbolic distance.

\textbf{M$\ddot{o}$bius addition.}
Direct adding two points within the Poinca\'{e} ball can result in a point that lies outside the boundary. Instead, M$\ddot{o}$bius addition~\cite{ganea2018hyperbolic} offers a counterpart to Euclidean addition specifically designed for hyperbolic space. The $M\ddot{o}bius 
\; addition$ of $\bm{x}$ and $\bm{y}$ in $\mathbb{B}^{d,\mathfrak{c}}$ is defined as:
 \begin{equation}
     \bm{x}\oplus_\mathfrak{c} \bm{y} = \frac{(1 + 2\mathfrak{c}\bm{x}^T\bm{y}+\mathfrak{c}\Vert \bm{y}\Vert ^2)\bm{x} + (1-\mathfrak{c}\Vert \bm{x}\Vert ^2)\bm{y}}{1+2\mathfrak{c}\bm{x}^T\bm{y}+\mathfrak{c}^2\Vert \bm{x}\Vert ^2\Vert \bm{y}\Vert ^2},
\end{equation}
 without loss of generality, the case $\mathfrak{c}>0$ can be reduced to $\mathfrak{c}=1$. This operation is neither commutative nor associative but it satisfies $\bm{x}\oplus_\mathfrak{c}\textbf{0}=\textbf{0}\oplus_\mathfrak{c}\bm{x}=\bm{x}$. When $\mathfrak{c}=0$, it reduces to the addition of vectors in Euclidean space.

\textbf{Hyperbolic Distance.} Distance in hyperbolic space is governed by metrics reflecting its curvature. The distance between two points $\bm{x}$ , $\bm{y}$ in $\mathbb{B}^{d,\mathfrak{c}}$ is defined as:
\begin{equation}
 d(\bm{x},\bm{y}) = \frac{2}{\sqrt{\mathfrak{c}}}\mathrm{arctanh}(\sqrt{\mathfrak{c}}\Vert -\bm{x}\oplus_\mathfrak{c}\bm{y}\Vert ),
 \end{equation}
 when $\mathfrak{c}=0$, $d(\bm{x},\bm{y})\rightarrow 2\Vert \bm{x} - \bm{y}\Vert $, recovers to Euclidean geometry. And when $\mathfrak{c}=1$, we can get the induced distance function:
\begin{equation}
d(\bm{x}, \bm{y})=\mathrm{arcosh}(1+2\frac{\Vert \bm{x}-\bm{y}\Vert ^2}{(1-\Vert \bm{x}\Vert ^2)(1-\Vert \bm{y}\Vert ^2)}).
 \end{equation}
 
The variable resolution of hyperbolic geometry is demonstrated by the exponential increase in hyperbolic distances as one moves away from the origin. This property allows hyperbolic space to naturally represent hierarchical or tree-like structures.

%% file: 4_methodology.tex
\section{METHODOLOGY}

\begin{figure*}
    \centering
    
    \includegraphics[width=\linewidth]{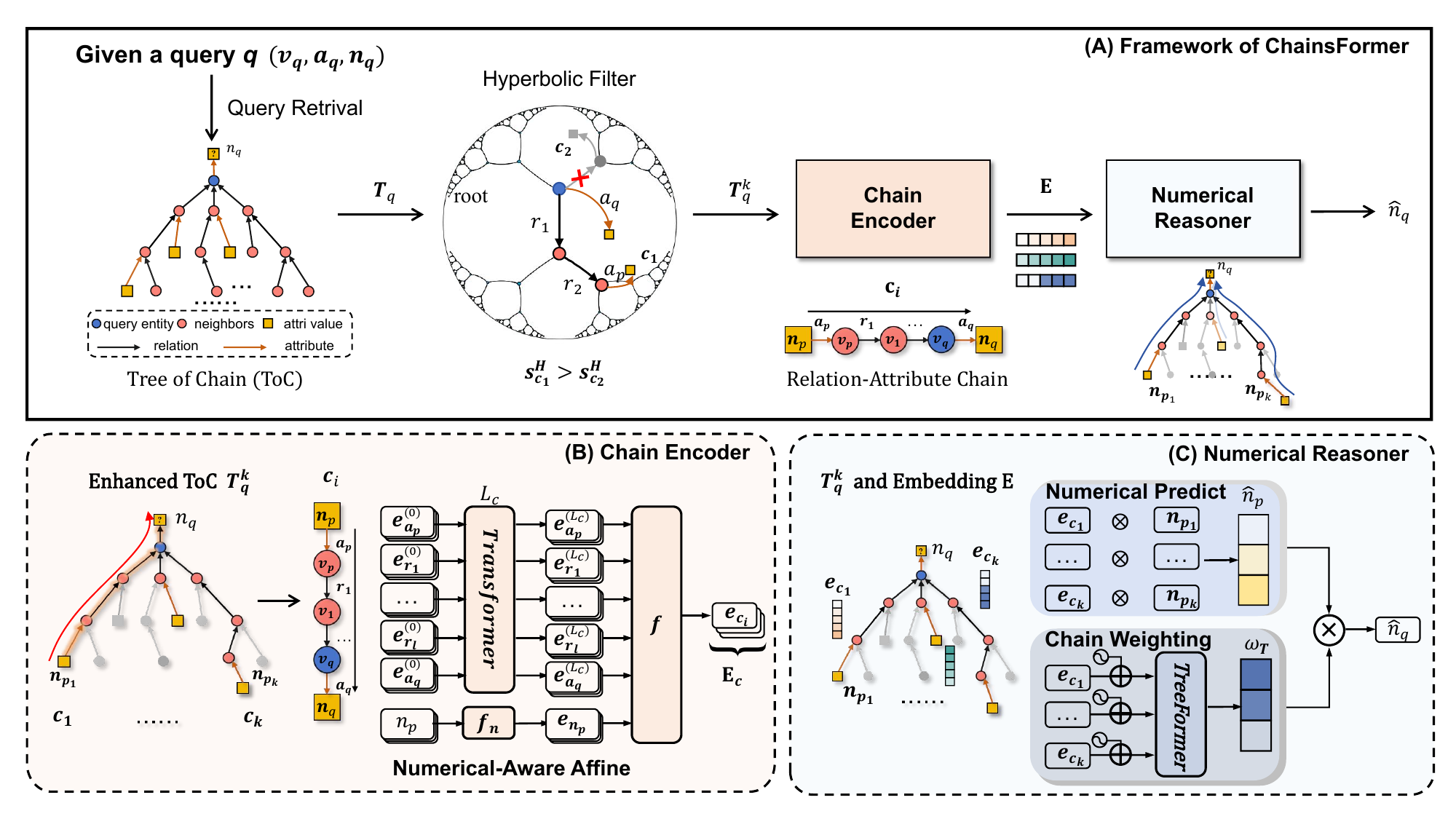}
    \caption{The overall architecture of ChainsFormer is shown in (A). ChainsFormer has for key components: Query Retrieval, Hyperbolic Filter, Chain Encoder and Numerical Reasoner. The Chain Encoder, detailed in (B), employs Transformer to represent RA-Chains. The Numerical Reasoner, detailed in (C), utilizes the proposed attention mechanism to weight and reason over the chains.}
    \label{fig:system}
    \vspace{-0.55cm}
\end{figure*}

In this section, we provide a detailed description of ChainsFormer. First, we explain the construction of Relation Attribute Chain. Next, we describe four key components of ChainsFormer, as shown in Figure~\ref{fig:system}: Query Retrieval, Hyperbolic Filter, Chain Encoder and Numerical Reasoner. For a given query, the process unfolds as follows: (1) Query Retrieval: Candidate chains are identified through a search process. (2) Hyperbolic Filter: Relevant chains are refined and filtered. (3) Chain Encoder: The selected chains are encoded into hidden representation domain. (4) Numerical Reasoner: Numerical inference is performed to derive the final results.

\subsection{From Graph to Chain Reasoning}

Both KGE-based and GNN-based methods introduce significant redundancy when aggregating information, losing the ability to identify relevant reasoning paths among various paths. Inspired by the step-by-step reasoning paradigm of Chain-of-Thought (CoT), we transform numerical reasoning from graph-based reasoning to chain-based reasoning. 

Specifically, we perform numerical reasoning by learning reasoning patterns for each logic chain. For example, for any known numerical triples, e.g., $(v_p, a_p, n_p)$, and the query $q = (v_q, a_q, ?)$ for $n_q$, a complete logic chain for numerical reasoning from $n_p$ to $n_q$ can be represented as $n_p \xrightarrow{a_p} v_p \xrightarrow{r_1} v_1 \xrightarrow{r_2} \dots \xrightarrow{r_l} v_q \xrightarrow{a_q} n_q$, where $v_l \in \mathcal{V}$ denotes $l$-hop neighbors of $v_p$, $r \in \mathcal{R}$ denotes relations that connect entities.

\textbf{Relation Attribute Chain (RA-Chain).} In numerical reasoning, reasoning patterns are primarily determined by the relations and attributes within chain, while the specific entities themselves are not critical for the reason patterns. Based on this observation, the reasoning patterns in this paper take the conjunctive form:
\begin{equation}
\begin{aligned}
R_h(n_p, n_q) &\gets R_{b_0}(n_p, a_p) \land R_{b_1}(a_p, r_1) \land \dots \\
&\qquad \land R_{b_l}(r_l, a_q) \land R_{b_{l+1}}(a_q, n_q),
\end{aligned}
\end{equation}
where $R_h$ is called rule head and $R_{b_0}(n_p, a_p)\land \dots \land R_{b_{l+1}}(a_q, n_q)$ is called rule body. Further, we tokenize the logic chain into a Relation-Attribute Chain (RA-Chain), represented as an ordered sequence of attributes and relations:
\begin{equation}
    c = (a_p, r_1, r_2, ...r_l, a_q).
\end{equation}

The construction of RA-Chain profiles the reasoning patterns and explicitly aligns to the step-by-step reasoning process for numerical attributes. enabling precise and explicit numerical reasoning. Moreover, as the length increases, it flexibly extends the reasoning depth to multiple hops.

\subsection{Query Retrieval}

For each query $q$, there can be an enormous number of noisy RA-Chains, potentially reaching hundreds of thousands or even millions. To reduce computational costs, ChainsFormer performs query-guided retrieval in the KG to construct a \textbf{Tree of Chain (ToC)} for sampling and training. This retrieval process involves a sufficient number of random walks (denoted as $N_s$) to search for RA-Chains. During retrieval, each chain $c_i$ is paired with its associated numerical attribute $n_{p_i}$, ensuring that the ToC captures both logical structures and numerical information. To avoid redundancy, cycles encountered during the walks are removed. The retrieved RA-Chains and their corresponding numerical attributes are then aggregated into the ToC, represented as:
\begin{equation}
    \mathcal{T}_q = \bigcup_{i=1}^{N_s}\{(c_i, n_{p_i})\},
\end{equation}
where $c_i$ denotes the $i$-th RA-Chain retrieved and $n_{p_i}$ represents its corresponding numerical attribute. 

\subsection{Hyperbolic Filter}

Given a query $q=(v_q, a_q, n_q)$, the number of possible reasoning paths becomes immense when all chains in ToC are considered, introducing both logic noise and high computational costs. Hyperbolic space, with its variable resolution, offers an efficient structure for representing such tree-like structures, making it promising for filtering irrelevant RA-Chains. Our findings indicate that low-dimensional hyperbolic filtering performs nearly as effectively as high-dimensional Euclidean filtering, motivating its adoption for RA-Chains. To this end, we propose the Hyperbolic Filter, which filters out noisy RA-Chains and enhance ToC through two key modules: hyperbolic chain embedding and hyperbolic affinity scoring.

\textbf{Hyperbolic Chain Embedding.} Inspired by translation-based embedding models~\cite{transe} and relative relation distances~\cite{NBFNet}, we propose a simple Hyperbolic Chain Embedding to realize the embedding of RA-Chains in hyperbolic space. Specifically, for any RA-Chain $c=(a_p,r_1,r_2,...,r_l,a_q)$, we initialize the hyperbolic representations $h_a$ and $h_r$ of attribute $a$ and relation $r$, and use M$\ddot{o}$bius addition to obtain the relational chain representations for c:
\begin{equation}
    \bm{h}_{c} = \bm{h}_{r_1} \oplus \bm{h}_{r_2} \oplus ... \oplus \bm{h}_{r_l},
\end{equation}
where $\bm{h}_{r_l}\in \mathbb{B}^{d,\mathfrak{c}}$ denotes the embedding of $r_l\in \mathcal{R}$ in $d$ dimensional Poincar\'{e} ball. $\bm{h}_{c}\in \mathbb{B}^{d}$ encapsulates logic length and various types of relations in the RA-Chain. This process can be viewed as a translation of relations in hyperbolic space, providing a simple but efficient representation of massive RA-Chains.

\textbf{Hyperbolic Affinity Score.} To select relevant RA-Chains and enhance the ToC for numerical reasoning, we propose a Hyperbolic Affinity Score $s_c^{H}$. This score assesses the effectiveness of the RA-Chain $c$  from inter and intra perspectives. The inter-score measures the relevance between the relations $(r_1, r_2, ... , r_l)$ and query attribute $a_q$. And the intra-score evaluates the match level between the known attribute $a_p$ and the query attribute $a_q$. 

The inter-score is computed by measuring the hyperbolic distance between the query attribute $a_q$ and the relational chain:

\begin{equation}
    d(\bm{h}_c, \bm{h}_{a_q}) = \mathrm{arcosh}(1+2\frac{\Vert \bm{h}_{a_q}-\bm{h}_c\Vert ^2}{(1-\Vert \bm{h}_{a_q}\Vert ^2)(1-\Vert \bm{h}_c\Vert ^2)}),
\end{equation}
where $\bm{h}_a \in \mathbb{B}^{d,\mathfrak{c}}$ denote  hyperbolic embedding of attribute. 

The intra-score, on the other hand, evaluates the similarity between the query attribute $a_q$ and the known attribute $a_p$, computed using hyperbolic distance to obtain $d(\bm{h}_{a_p}, \bm{h}_{a_q})$.

The hyperbolic affinity score can be obtained by the following equation:
\begin{equation}
     s_c^{H} = \lambda d(\bm{h}_{a_p}, \bm{h}_{a_q}) + (1-\lambda) d(\bm{h}_c, \bm{h}_{a_q}),
\end{equation}
where $\lambda$ is a hyperparameter that balances the inter- and intra-scores.

For each query $q$ and its corresponding ToC $\mathcal{T}_q$, the Hyperbolic Filter computes the hyperbolic affinity score for all RA-Chains and selects the top-$k$ candidates to form the Enhanced ToC, denotes as $\mathcal{T}_q^k$:

\begin{equation} \mathcal{T}_q^k = \{ c_i \in \mathcal{T}_q \mid \text{card}(\{ c_j \in \mathcal{T}_q \mid s_{c_j}^{H} > s_{c_i}^{H} \}) < k \}, \end{equation}
where $\text{card}$ denotes the cardinality of the set. This process reduces the search space and retains most relevant RA-Chains, enabling more efficient and accurate numerical reasoning.

\subsection{Chain Encoder}

For a query $q$ and the Enhanced ToC $\mathcal{T}_q^k$, we propose a Chain Encoder to encode each RA-Chain within it. The Chain Encoder consists of two components: In-Context Chain Representation, which employs a Transformer-based sequence encoder to model RA-Chains, enabling in-context representation for constructing a step-by-step reasoning framework, and Numerical-Aware Affine Transfer, which integrates attributes with varying distributions into chain embeddings, allowing RA-Chains to adaptively perceive numerical values.

\textbf{In-Context Chain Representation.}
Given a query $q=(v_q, a_q, n_q)$ and a RA-Chain $c = (a_p, r_1, r_2, ..., r_l, a_q)$, we encode the chain using an encoder-only Transformer model. By tokenizing and representing the sequential structure of the chain, ChainsFormer captures the in-context associations between relations and attributes, effectively modeling step-by-step reasoning patterns. The input to Transformer $\bm{P}^{(0)}$ is:
\begin{equation}
\bm{P}^{(0)} = [\bm{e}_{a_p}^{(0)}\Vert \bm{e}_{r_l}^{(0)}\Vert ...\Vert \bm{e}_{r_1}^{(0)}\Vert \bm{e}_{a_q}^{(0)}\Vert \bm{e}_{end}^{(0)}],
\end{equation}
\begin{equation}
    \bm{e}_{r_l}^{(0)} = \mathrm{log}(\bm{h}_{r_l}) = \mathrm{arctanh}(\Vert \bm{h}_{r_l}\Vert)\frac{\bm{h}_{r_l}}{\Vert \bm{h}_{r_l} \Vert},
    \label{equation 7}
\end{equation}
where $\bm{e}_{a}^{(0)}, \bm{e}_{r}^{(0)}\in \mathbb{R}^{d}$ denote the projection vector of attribute $a$ and relation $r$ in Euclidean space, which can be get by equation~\ref{equation 7}, we use logarithmic-map projects the vector from $\mathbb{B}^{d,\mathfrak{c}}$ to $\mathbb{R}^{d}.$ The $\bm{e}_{end}^{(0)}$ denotes a common end token embedding for each logic chain. $[ \bm{e}_1\Vert ... \Vert\bm{e}_n]$ is a horizontal concatenation of vector. 

We then learn the interdependencies of relations and attributes via the attention layer:
\begin{equation}
\tilde{\bm{P}}^{(i)} = \bm{V}^{(i)}\bm{P}^{(i)}softmax(\frac{(\bm{Q}^{(i)}\bm{P}^{(i)})^T (\bm{K^{(i)}P^{(i)})}}{\sqrt{d}}),
\end{equation}
where $\bm{Q}^{(i)}, \bm{K}^{(i)}, \bm{V}^{(i)}\in \mathbb{R}^{d\times d}$ are projection matrices for query, key and value respectively. We apply $L_c$ layers of Transformer, utilizing multi-head attention mechanism~\cite{attention}, residual connection~\cite{resnet}, followed by layer normalization. By repeating the above process for $i = 0,..., L_c-1,$ we get the final representation $\bm{P}^{(L_c)}$ and select $\bm{e}_{end}^{(L_c)}$ as the chain representation, denotes as $\bm{e}_c$, which contains the complete step-by-step reasoning information from $a_p$ to $a_q$.

\textbf{Numerical-Aware Affine Transfer.} The magnitude of the numerical value $n_p$ associated with the RA-Chain also affects the reasoning results. However, due to large differences in the numerical distributions across attributes (e.g., a height of 1.81m versus a population of 3.1e9), directly integrating these numerical features into the logic chain is challenging~\cite{PLM, nlp_survey}. Therefore, we design the Numerical-Aware Affine Transfer, which adaptively generates numerical parameters via Affine network for value-aware numerical reasoning.

To simplify the explanation, we select a RA-Chain $c$, the chain representation $e_c$ get by Transformer and the connected attribute value $n_p$. Since the vast numerical range and severe sparsity in value distribution make numerical embedding challenging, we first encode the continuous value $n_p$ into a form more easily understood by machines. Specifically, we map $n_p$ from the real-value space to a Float64 0-1 bit-stream:
\begin{equation}
\bm{e}_{n_p} = f_n(n_p),
\end{equation}
where$f_{n}:\mathbb{R}\rightarrow\mathbb{R}^{64}.$ This approach draws inspiration by number embedding in natural language processing that utilize exponential or scientific notation to render numerical values more comprehensible for machine processing~\cite{nlp_survey}.

Then, we generate numerical parameters $E_{n_p}= [E_{n_p}^\alpha, E_{n_p}^\beta]$ based on attribute value $n_p$:
\begin{equation}
    E_{n_p}^\alpha = \text{MLP}_\alpha(\bm{e}_{n_p}; \phi_\alpha), E_{n_p}^\beta = \text{MLP}_\beta(\bm{e}_{n_p}; \phi_\beta),
\end{equation}
where $E_{n_p}^\alpha\in \mathbb{R}^{d\times d}$ denotes rotation matrix, $E_{n_p}^\beta\in \mathbb{R}^{d}$ denotes bias vector. Afterwards, we perform affine projection~\cite{oxygenerator} on the latent embedding of the logic chain to perform numerical value-aware feature scaling:

\begin{equation}
    \bm{\tilde{e}}_{c} = \bm{e}_c \otimes E_{n_p} \triangleq E_{n_p}^{\alpha^T} \cdot \bm{e}_c + E_{n_p}^{\beta},
\end{equation}
where $\bm{e}_c$ is the output of the Transformer, $\bm{\tilde{e}}_{c}$ is the value-aware adaptive representation for RA-Chain $c$. In this way, we apply affine transformation to each RA-Chain based on its numerical magnitude, achieving effective representation and utilization of numerical features across chains.

Chain Encoder computes the value-aware chain representation $\bm{\tilde{e}}_{c}$ for each RA-Chain $c_i\in\mathcal{T}_q^k$, concatenating them to form the chain representation matrix $\mathrm{\bm{E}} = [\bm{{\tilde{e}}}_{c_1} \Vert ... \Vert {\bm{\tilde{e}}}_{c_i} \Vert ... \Vert {\bm{\tilde{e}}}_{c_k}]$.

\subsection{Numerical Reasoner} 

Given a query $q$ and its Enhanced ToC $\mathcal{T}_q^k$, the Numerical Reasoner, as shown in Figure~3 (C), is designed to perform Numerical Prediction for each RA-Chain and Chain Weighting to evaluate the contribution of each chain and identify the key reasoning paths.

\textbf{Numerical Prediction.}  For each chain $c$, the Numerical Reasoner models the reasoning process of inferring the query value $n_q$ from the known value $n_p$ along the chain. Specifically, this process is formulated as a transformation of numerical attribute values, represented by the general operation $n_q = n_p \oplus \bm{e}_c$. This operation, referred to as numerical projection along attributes and relations, captures the value shifts across the RA-Chain. We evaluate three different numerical projection methods to assess their effectiveness through experiments.

The first approach is Translation Projection in the numerical domain, this method directly adds a translation factor $\beta$ to the known value $n_p$, representing how $n_p$ is translated along the chain:
\begin{equation}
    \hat{n}_q = n_p + \beta, \\
    \beta = MLP(\bm{\tilde{e}}_c;\theta_{MLP}),
\end{equation}
denotes the linear shift in numerical values.
The second approach involves scaling projection, which scales the known value $n_p$ by a factor $\alpha$, adapting the numerical value based on the logic's context:
\begin{equation}
    \hat{n}_q = \alpha * n_p, \alpha =  MLP(\bm{\tilde{e}}_c;\theta_{MLP}),
\end{equation}
which modifies the magnitude of $n_p$ according to the characteristics of the numerical neural logic chain.
The third approach combines both:
\begin{equation}
\hat{n}_q = \alpha * (n_p + \beta), \;\;\;\;\alpha, \beta =  MLP(\bm{\tilde{e}}_c;\theta_{MLP}),
\end{equation}
the scaling projection is preferred for its versatility across various numerical ranges and its ability to enable reasoning across diverse attribute types.

\textbf{Logic Chain Weighting.} Since the Enhanced ToC $\mathcal{T}_q^k$ contains a variety of RA-Chains, the interactions between these heterogeneous paths provide additional insights for reasoning. Therefore, we employ another Treeformer to evaluate the relative importance of each chain and to identify key RA-Chains. By leveraging the distinct contributions of each chain, this approach allows for traceability of reasoning paths, thereby enhancing the transparency of the reasoning outcomes.

The input of Treeformer $\textbf{C}^{(0)}$ is:
\begin{equation}
\textbf{C}^{(0)} = |(\bm{\tilde{e}}_{c_1}\Vert \bm{\tilde{e}}_{c_2}\Vert ...\Vert \bm{\tilde{e}}_{c_k})| + f_{len}(c_1, ..., c_k).
\end{equation}
where $\bm{\tilde{e}}_p \in \mathbb{R}^d$ denotes the chain representation obtained by Chain Encoder, and $f_{Len} \in \mathbb{R}^d$ is a learnable layer for length encoding. Positional encoding is omitted as the order of logic chain is not crucial. Instead, we incorporate length encoding to represent the order of RA-Chain.

The Tree representation $\textbf{C}^{(L_c)}$ after $L_c$ layers reflects the influence and relative importance of different RA-Chains in Enhanced ToC $\mathcal{T}_q^k$ for numerical reasoning. The output then passes through a linear layer with softmax:

\begin{equation}
\bm{\omega}=\mathrm{softmax}(MLP_t(\textbf{T}^{(L_c)}); \phi_t),
\end{equation}
$\bm{\omega}\in\mathbb{R}^{k},$ which reflects the important scores of each chain. A higher $\omega$ score indicates a RA-Chain is more likely to influence reasoning.

By characterizing the importance of different RA-Chains, Numerical Reasoner simulates human decision-making. It assesses the predictive power of each information source and weights them accordingly. Finally, we weight and sum the chain predictions using the importance scores:
\begin{equation}
\hat{n}_q = \sum_{i=1}^k \omega_{c_i} \hat{n}_{p_i},
\end{equation}
where $\omega_{c_i}$ denotes the importance score of $c_i$, $\hat{n}_q$ denotes the the predicted value of $n_q$.

\subsection{Training and Optimization.}

\textbf{Loss Function.} To optimize numerical reasoning for ChainsFormer, we define a loss function to measure the discrepancy between the predicted numerical value $\hat{n}_q$ and the actual value $n_q$. Since the  differing ranges of numerical attributes, we apply min-max normalization to eliminate this bias before calculating the loss. The min-max normalization of a value $n$ is defined as:
\begin{equation}
    norm(n_q) = \frac{n_q-min(n_a)}{max(n_a)-min(n_a)},
\end{equation}
where $\min(n_a)$ and $\max(n_a)$ are the minimum and maximum values of attribute $a$. We define the following loss function $\mathcal{L}$ of our model to minimize the prediction error:

\begin{equation}
    \mathcal{L} = \frac{1}{|\mathcal{E}_a|}\sum_{(v_q, a_q, n_q)\in \mathcal{E}_a}(norm(n_q)-norm(\hat{n}_q))^2,
\end{equation}
where $norm(\hat{n}_q)$ denotes the normalized prediction result, $\mathcal{L}$ is the mean squared error loss.

\textbf{Model Training Process.} In each epoch, for every query $q = (v_q, a_q, n_q)$, we first perform query-guided retrieval to construct the Tree of Chain $\mathcal{T}_q$, which contains diverse RA-Chains. The ToC is then filtered in hyperbolic space using the Hyperbolic Filter to extract relevant chains, forming $\mathcal{T}_q^k$. Next, the Chain Encoder is employed to represent the chains, followed by numerical reasoning and result traceability through the Numerical Reasoner, enabling precise and interpretable numerical reasoning in knowledge graphs. The detailed training procedure of ChainsFormer is summarized in Algorithm \ref{alg:training_procedure}.

\begin{algorithm}
\caption{Training Procedure for ChainsFormer} \label{alg:training_procedure}
\begin{algorithmic}[1]
\REQUIRE Query set $Q_{query}$ with missing attribute values, convergence threshold $\epsilon$.
\ENSURE Predicted results and chain scores for each query.
\STATE Initialize embeddings for relations, attributes, and model parameters.
\STATE Initialize previous total loss $\mathcal{L}_{prev} = +\infty$.
\REPEAT
    \STATE Set total loss $\mathcal{L} = 0$.
    \FOR{each query $(v_q, a_q, n_q) \in Q_{query}$}
        \STATE Perform query-guided retrieval to construct the ToC $\mathcal{T}_q$, containing diverse RA-Chains.
        \STATE Filter chains using the Hyperbolic Filter to obtain Enhanced ToC $\mathcal{T}_q^k$ (Eq.~7-Eq.~10).
        \STATE Encode chains using the Chain Encoder to obtain chain embeddings (Eq.~11-Eq.~16).
        \STATE Perform numerical reasoning along each chain using the Numerical Reasoner (Eq.~17-Eq.~19).
        \STATE Calculate chain contribution scores through Chain Weighting to identify critical paths (Eq.~20-Eq.~21).
        \STATE Aggregate chain predictions to obtain the final prediction $\hat{n}_q$ and compute regression loss $\mathcal{L}_q$ (Eq.~22-Eq.~24).
        \STATE Accumulate the loss: $\mathcal{L} = \mathcal{L} + \mathcal{L}_q$.
    \ENDFOR
    \STATE Check convergence: $\Delta \mathcal{L} = |\mathcal{L}_{prev} - \mathcal{L}|$.
    \STATE Update $\mathcal{L}_{prev} = \mathcal{L}$.
    \STATE Update model parameters using accumulated loss $\mathcal{L}$.
\UNTIL{$\Delta \mathcal{L} < \epsilon$}
\end{algorithmic}
\end{algorithm}
\vspace{-3pt}
\subsection{Complexity Analysis}
We analyze the computational complexity of our proposed model. The time complexity is $\mathcal{O}(N_sd+kd^2)$, where $N_s$ denotes the number of random walks, $k$ denotes the filtering count, and $d$ denotes the hidden dimensions. Specifically, our algorithm consists of three main steps. First, the Query Retrieval and  Hyperbolic Filter involves random walks and the computation of hyperbolic affinity scores, resulting in a time complexity of $\mathcal{O}(N_s + N_sd) \approx \mathcal{O}(N_sd)$. Next, the Chain Encoder utilizes a Transformer-based architecture for in-context representations, with a complexity of $\mathcal{O}(k(d^2l+dl^2))$, where $l$ is the sequence length. Since $l$ is a small constant, the complexity of the Chain Encoder primarily depends on $\mathcal{O}(kd^2)$. Finally, the Numerical Reasoner calculates chain contributions, with the main complexity lying in path weighting, resulting in $\mathcal{O}(kd^2)$. Notably, the computational cost remains low as ChainsFormer avoids exhaustive traversal of the entire graph. Its sequence-based design also enables parallel processing, ensuring high efficiency for real-world applications.

%% file: 5_experiment.tex
\definecolor{lgray}{rgb}{0.9,0.9,0.9}
\section{EXPERIMENT}
\label{5_Experiment}

In this section, we evaluate the effectiveness of the proposed method and conduct extensive experiments. More comprehensive in-depth analysis are presented in detail with the aim of answering the following research questions.

\begin{itemize}
    \item \textbf{RQ1:} How does ChainsFormer perform against other baselines in numerical reasoning?
    \item \textbf{RQ2:} How does transitioning from graph to chains enhance the reasoning depth and capability?
    \item \textbf{RQ3:} How effective is each part of the proposed model?
    \item \textbf{RQ4:} What impact do hyperparameters and numerical projection methods have on numerical reasoning?
\end{itemize}
\subsection{Setup}

\textbf{Datasets.} We conduct experimental evaluation on two widely used datasets: FB15K-237~\cite{FB15K-237} and YAGO15K~\cite{YAGO15K}, all numerical attributes were introduced in \cite{MMKG}. These datasets have been commonly adopted, providing a solid basis for evaluation. They include a total of 16 numerical attribute types, categorized into temporal, spatial, and quantity classes. The datasets are relatively large in scale, and both are divided into train, validation, and test sets in an 8:1:1 ratio. Detailed statistics are presented in Table \ref{tab:dataset statistics} and Table \ref{tab:attribute statistics}.
\label{Appendix:dataset}
\begin{itemize}
    \item \textbf{FB15K-237:}  FB15K-237 is a subset of Freebase~\cite{FB}, primarily covering facts about movies,actors, sports and so on. FB15K-237 is derived from the benchmark FB15K by removing inverse relations. MMKG~\cite{MMKG} introduces a total of 116 numerical attributes, amounting to 29,220 entries. To ensure experimental fairness, we maintain consistency with KGA~\cite{KGA} by selecting 9 of these numerical attributes, totaling 23,154 entries, and using the same spilt with~\cite{KGA}.

    \item \textbf{YAGO15K:} YAGO15K is a subset of YAGO: Yet Another Freat Ontology. In addition to accounting for about 40\% of FB15K-237's data, it combines data from Wikipedia, WordNet, and GeoNames. MMKG~\cite{MMKG} introduces a total of 7 numerical attributes, amounting to 23520 entries. We select all numeric attributes and divide them in the same way with~\cite{KGA}.
\end{itemize}

\begin{table}[ht]
\centering
\caption{Statistics of the datasets used in this paper.}
\label{tab:dataset statistics}
\begin{tabular}{lccccc}
\toprule
 \textbf{Statistics}& \textbf{$|\mathcal{V}|$} & \textbf{$|\mathcal{R}|$} & \textbf{$|\mathcal{A}|$} &\textbf{$|\mathcal{E}_r|$} & \textbf{$|\mathcal{E}_a|$}  \\
\midrule
\textbf{YAGO15K} & 15,404 & 32 & 7 & 122,886 & 23,520\\
\textbf{FB15K-237} & 14,951 & 237 & 11 & 310,116 & 23,154\\
\bottomrule
\vspace{-10pt}
\end{tabular}
\end{table}

\begin{table}[ht]
\centering
\caption{Statistics of the numerical attributes in two datasets.}
\label{tab:attribute statistics}
\begin{tabular}{lccccc}
\toprule
 \multicolumn{2}{c}{\textbf{Statistics}}&\textbf{$|\mathcal{E}_a|$} & \textbf{min(a)} &\textbf{max(a)} &\textbf{max-min}\\
\midrule
\multirow{8}{*}{\textbf{YG}} & birth  & 8217 & 354.9 & 2014.0&1659.1\\
& death & 1821 & 348.0 & 2161.1& 1813.1\\
& created & 6576 & 100.0 & 2018.7& 1918.7\\
& destroyed &537  & 476.0 & 2017.2&1541.2 \\
& happened & 388 & 218.0 &2018.2 &1800.2 \\
& latitude & 2989 & -51.7 &73.0 &124.7\\
& longitude & 2989 & -175.0 & 179.0&354.0\\
\midrule
\multirow{10}{*}{\textbf{FB}} & birth & 4406 & -383.0 & 1999.9&2382.9\\
& death & 1214 & -322.0 &2015.6 &2337.6 \\
& film\_release & 1853 & 1927.1 &2013.5 &86.4 \\
& org\_founded & 1228 & 1088.0 &2013.0 &925.0 \\
& loc\_founded & 917 &-2999.0  &2011.6 &5010.6\\
& latitude & 3190 & -90.0 &77.6 &167.6 \\
& longitude & 3192 & -175.2 & 179.2& 354.4\\
& area & 2154 & 1.0 & 1.7e8&1.7e8 \\
& population & 1920 & 1.0 & 3.1e9& 3.1e9\\
& height & 2855 & 1.34 &2.18 &0.84 \\
& weight & 225 & 44.0 &147.0 & 103.0\\
\bottomrule
\end{tabular}
\vspace{-12pt}
\end{table}

\begin{table*}[htbp]
\centering
\caption{Performance comparison of numerical reasoning on two datasets. In each column, the best results are highlighted in bold and shaded. To evaluate the overall model performance, we normalize all attributes to a range of 0-1 and compute the average MSE and RMSE across all normalized classes ,denotes as $\textbf{Average}^*$.}
\resizebox{1\textwidth}{!}{
\label{overall performance}
\begin{tabular}{lrrrrrrrr|rrrrrrr}
\toprule
\multicolumn{2}{c}{\multirow{2}{*}{\textbf{Attribute}}} & \multicolumn{7}{c}{\textbf{RMSE}} & \multicolumn{7}{c}{\textbf{MAE}} \\
\cmidrule(lr){3-9} \cmidrule(lr){10-16}
& & \textbf{NAP++} & \textbf{MrAP} & \textbf{PLM-reg} & \textbf{KGA} & \textbf{HyNT} & \textbf{ToG-R} &\textbf{Ours} & \textbf{NAP++} & \textbf{MrAP} & \textbf{PLM-reg} & \textbf{KGA} & \textbf{HyNT} & \textbf{ToG-R} &\textbf{Ours} \\
\midrule
\multirow{8}{*}{\textbf{YG}} & death & 99.4 & 84.2 & 146.6 & 85.6 & 40.5 & 190.4& \cellcolor{lgray}\textbf{20.7} & 45.7 & 34.0 & 49.2 & 32.0 & 20.7&115.2 & \cellcolor{lgray}\textbf{15.5} \\
& birth & 59.9 & 31.5 & 36.4 & 57.4 & \cellcolor{lgray}\textbf{29.5} &160.1& 56.7 & 23.2 & 19.7 & 18.8 & 16.3 & 19.1 &81.5& \cellcolor{lgray}\textbf{15.8} \\
& created & 152.3 & 149.6 & 133.5 & 134.2 & 142.6 & 204.1& \cellcolor{lgray}\textbf{131.4} & 83.5 & 70.4 & 62.1 & 60.4 & 73.6& 128.9 & \cellcolor{lgray}\textbf{57.4} \\
& destroyed & 75.5 & 62.0 & 45.7 & 39.2 & 52.7 &95.3& \cellcolor{lgray}\textbf{38.3} & 38.2 & 34.6 & 29.8 & 26.0 & 36.6&59.6 & \cellcolor{lgray}\textbf{25.1} \\
& happened & 159.9 & 73.8 & 51.9 & 63.7 & 47.7&184.6 & \cellcolor{lgray}\textbf{39.6} & 73.7 & 54.1 & 38.4 & 36.0 & 32.5 & 111.4&\cellcolor{lgray}\textbf{26.5} \\
& latitude & 13.8 & 7.9 & 7.6 & 8.5 & 15.0 &6.7& \cellcolor{lgray}\textbf{6.6} & 8.7 & 2.8 & 4.2 & 3.5 & 7.4 &2.3& \cellcolor{lgray}\textbf{2.2} \\
& longitude & 58.6 & 17.1 & 29.6 & 23.2 & 34.6 &17.8& \cellcolor{lgray}\textbf{15.1} & 43.1 & 5.7 & 17.0 & 8.0 & 17.5 &4.4& \cellcolor{lgray}\textbf{3.7} \\
& $\textbf{Average}^*$ & 0.087 & 0.052 & 0.056 & 0.054 & 0.068 &0.089& \cellcolor{lgray}\textbf{0.047} & 0.044 & 0.021 & 0.025 & 0.020 & 0.029 & 0.045&\cellcolor{lgray}\textbf{0.016} \\
\midrule
\multirow{10}{*}{\textbf{FB}} & birth & 34.3 & 38.6 & 39.1 & 111.4 & 34.9 & 71.5& \cellcolor{lgray}\textbf{27.4} & 22.1 & 15.6 & 25.0 & 18.9 & 17.8 &29.4& \cellcolor{lgray}\textbf{14.4} \\
& death & 85.2 & 32.2 & 72.6 & 37.8 & 63.0 & 74.9 & \cellcolor{lgray}\textbf{31.1} & 53.2 & \cellcolor{lgray}\textbf{16.3} & 46.9 & 20.6 & 34.5 &38.7& 19.4 \\
& film\_release & 14.7 & 8.6 & 11.9 & \cellcolor{lgray}\textbf{5.1} & 10.5 &18.1 &9.1 & 9.9 & 6.3 & 5.8 & \cellcolor{lgray}\textbf{4.0} & 8.1 & 14.1& 5.1 \\
& org\_found & 98.2 & 93.1 & 101.5 & 100.7 & 105.3&94.9 & \cellcolor{lgray}\textbf{91.3} & 59.3 & 59.2 & 66.8 & 54.5 & 57.9 &60.5& \cellcolor{lgray}\textbf{53.3} \\
& loc\_founded & 277.0 & \cellcolor{lgray}\textbf{158.6} & 214.8 & 171.9 & 245.5 & 355.2&205.1 & 149.9 & 103.7 & 145.3 & \cellcolor{lgray}\textbf{76.0} & 114.2 & 149.4& 99.4 \\
& latitude & 18.9 & 3.6 & 9.3 & 6.8 & 9.6&5.5 & \cellcolor{lgray}\textbf{3.3} & 11.8 & 1.5 & 5.6 & 2.9 & 5.1 &2.3& \cellcolor{lgray}\textbf{1.4} \\
& longitude & 71.8 & 9.7 & 25.2 & 19.0 & 28.6 &31.8& \cellcolor{lgray}\textbf{9.4} & 54.7 & 4.0 & 16.3 & 6.3 & 12.2&12.6 & \cellcolor{lgray}\textbf{3.6} \\
& area & 1.2e6 & 1.3e6 & 5.6e6 & 1.8e5 & 5.8e5 &6.1e6& \cellcolor{lgray}\textbf{1.4e5} & 4.4e5 & 5.3e5 & 1.5e6 & 4.8e4 & 5.0e5 &2.8e6& \cellcolor{lgray}\textbf{4.0e4} \\
& population & 6.5e7 & 4.3e7 & 2.3e7 & 1.7e7 & 1.8e7 &1.6e7& \cellcolor{lgray}\textbf{1.8e8} & 7.5e6 & 2.1e7 & 1.6e7 & 4.0e6 & 7.6e6 &6.1e7& \cellcolor{lgray}\textbf{3.9e6} \\
& height & 0.102 & 0.106 & 0.280 & 0.094 & 0.089 & 0.115&\cellcolor{lgray}\textbf{0.087} & 0.080 & 0.086 & 0.188 & 0.073 & 0.070 &0.090 &\cellcolor{lgray}\textbf{0.064} \\
& weight & 18.9 & 18.3 & 18.0 & 18.8 & 18.6 &24.4& \cellcolor{lgray}\textbf{17.1} & 15.9 & 12.9 & 11.3 & 13.6 & 15.1 &14.3& \cellcolor{lgray}\textbf{10.9} \\
& $\textbf{Average}^*$ & 0.114 & 0.063 & 0.129 & 0.062 & 0.66&0.088 & \cellcolor{lgray}\textbf{0.058} & 0.063 & 0.029 & 0.054 & 0.027 & 0.031 & 0.041&\cellcolor{lgray}\textbf{0.025} \\
\bottomrule
\end{tabular}
}
\vspace{-0.35cm}
\end{table*}

\textbf{Baselines.} In the experiment, we compare with five models. Here we introduce each baseline model in short.
\begin{itemize}
    \item \textbf{NAP++}~\cite{NAP++}: NAP++ selects the nearest k neighbors using TransE~\cite{transe} embedding and aggregates their numerical attributes.
    \item \textbf{MrAP}~\cite{MrAP}: MrAP defines propagation paths and performs message passing based on different relations.
    \item \textbf{PLM-reg}~\cite{PLM}: PLM-reg utilizes pre-trained language model representations for direct numerical prediction.
    \item \textbf{KGA}~\cite{KGA}: KGA buckets numbers into discrete entities and uses link prediction tasks for numerical reasoning.
    \item \textbf{HyNT}~\cite{HyNT}: HyNT treats numerical attributes as qualifiers of triples and combines entity representations for regression.
    \item \textbf{ToG}~\cite{ToG}: Think on Graph(ToG) is a new LLM-KG
integration paradigm which exploits the reasoning capability of LLMs for knowledge graphs, enabling search and pruning over KGs to complete various tasks.
\end{itemize}

\textbf{Implementation Details.}
\label{Appendix:Implementation Details}
We train the model for 200 epochs with early stopping, using the Adam optimizer (learning rate: 0.0001) and L1 loss. The average path lengths of FB15K-237 and YAGO15K are 2.72 and 2.99, so we set the random walk order $l=3$, with $N_s=2048$ and top-k selection of 256. The hidden layer dimensions are 256/128, and both the Chain Encoder and Numerical Reasoner use 2 Transformer layers with 4 attention heads. We implement the ChainsFormer based on PyTorch 1.8.1 framework. All the evaluated models are implemented on a server with two CPUs (Intel Xeon E5-2630) and four GPUs (NVIDIA GTX 4090, 24GB memory).

\textbf{Evaluation Metrics.} Due to the wide variation of numerical ranges, we primarily report the mean absolute error (MAE) for each kind of attribute. Additionally, the root mean square error (RMSE) is provided as a reference for evaluation. To evaluate the overall model performance, we normalize all attributes to a range of 0-1 and compute the average MSE and RMSE across all normalized classes.

\subsection{Performance Comparison (\textit{RQ1})}

We run ChainsFormer and other baselines and report the average test accuracy over all attributes in Table \ref{overall performance}. From the comprehensive views, we make the following observations:

Firstly, ChainsFormer significantly outperforms other baselines, ranking first in average performance, demonstrating the superiority of our method. 
It achieves MAE improvements of 20.0\% on YAGO15K and 7.4\% on FB15K-237. This performance boost indicates that ChainsFormer achieves stable improvement in numerical reasoning, providing more accurate predictions of numerical attributes.

Secondly, the improvement in spatial attributes (longitude and latitude) with ChainsFormer is particularly notable, achieving MAE improvements of 30.5\% and 9.1\% on two datasets. In contrast, MrAP and NAP++, which consider only one-hop neighbors, fail to capture the hidden connections within long distance numerical attributes, underscoring the importance of multi-hop information for numerical reasoning.

Thirdly, for quantity attributes, ChainsFormer outperforms in predicting area, population, height and weight in FB15K-237, achieving MAE improvement of 8.4\%. Unlike other models confined to specific ranges, ChainsFormer adapts robustly across a broad spectrum of quantity attributes, such as 3.1e9 in population and 0.84 in height, the detailed information given in Table~\ref{tab:attribute statistics}. The result demonstrates the effectiveness of Numerical-aware Affine Transfer.

For temporal attributes, ChainsFormer achieves significant improvements over the state-of-the-art model KGA, with improvements of 19.2\% and 3.65\% on YAGO15K and FB15K-237, respectively. On YAGO15K, ChainsFormer achieves the best performance across all temporal attributes for MAE. Experimental results demonstrate that ChainsFormer consistently delivers high performance for temporal attributes.

\begin{table}[h!]
\centering
\caption{Method comparison based on different reasoning capabilities.}
\label{tab:method_extensible_comparison}
\begin{tabular}{@{\hspace{0.1pt}}l|ccccc|c}
\hline
\hline
    & \textbf{NAP++} & \textbf{MrAP} & \textbf{PLM-reg} & \textbf{KGA} & \textbf{HyNT}& \textbf{Ours} \\ \hline
\textbf{Num-aware}              & \ding{55}              & \ding{55}      & \ding{55}          & \checkmark              & \checkmark     & \checkmark              \\ 
\textbf{One-hop}               & \checkmark             & \checkmark     & \checkmark           & \checkmark              & \checkmark   & \checkmark                \\ 
\textbf{Multi-hop}            & \ding{55}             & \ding{55}      & \ding{55}           & \checkmark              & \ding{55}  & \checkmark                 \\ 
\textbf{Same-attr}                & \checkmark              & \checkmark     & \checkmark          & \checkmark              & \checkmark   & \checkmark                 \\ 
\textbf{Multi-attr}               & \ding{55}              & \checkmark      & \ding{55}          & \ding{55}              & \checkmark  & \checkmark         \\ 
\hline
\hline
\end{tabular}
\vspace{-5pt}
\end{table}

\subsection{Effectiveness of Graph to Chain (\textit{RQ2})}
To investigate the effectiveness of the chain-based paradigm, we focus on two aspects: Enhanced Reasoning depth and Transparent, Step-by-Step Reasoning.

\begin{figure}[ht]
    \centering    
    \includegraphics[width=1.0\linewidth]{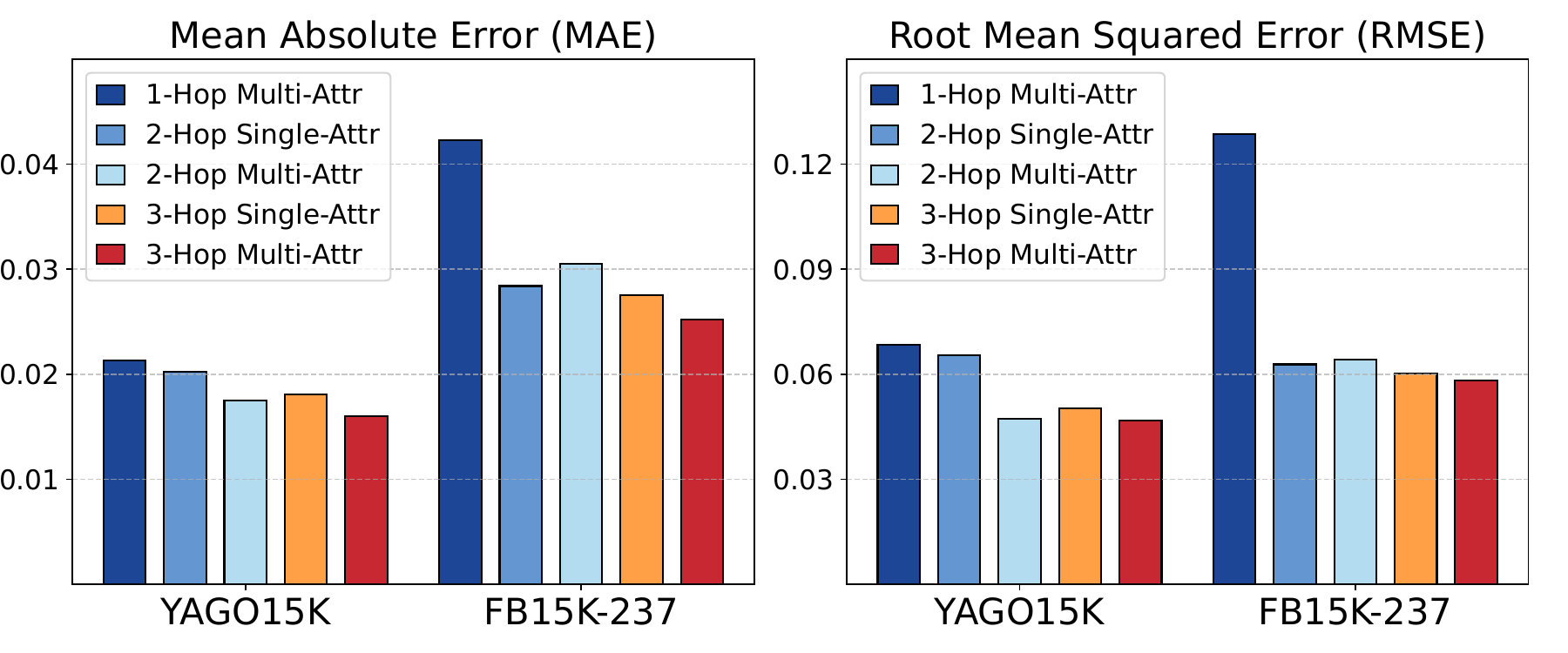}
    \vspace{-10pt}
    \caption{Performance comparison across different reasoning settings. The results are evaluated using Mean Absolute Error (MAE) and Root Mean Squared Error (RMSE) on the YAGO15K and FB15K-237 datasets. }
    \label{fig:extensible}
    \vspace{-15pt}
\end{figure}

\subsubsection{Enhanced Reasoning depth}
The graph to chain paradigm provides a new method to capture reasong pattern, allowing numerical reasoning capabilities to be extended. As shown in Table~\ref{tab:method_extensible_comparison}, existing methods either struggle to explicitly reason over multi-hop neighbors or fail to incorporate diverse attributes in reasoning tasks. ChainsFormer, on the other hand, extends reasoning depth to satisfy arbitrary multi-hop, multi-attribute using serialized reasoning chains. This chain-based approach enables ChainsFormer to fully exploit the information in the knowledge graph, leading to more accurate numerical reasoning.

\begin{table*}[ht]
\centering
\vspace{-10pt}
\caption{A case analysis for the most important RA-Chains identified by the Numerical Reasoner. We select random numerical attributes from both datasets to report.}
\label{table:key RA-Chain}
\begin{tabular}{lcl}
\toprule

\multicolumn{2}{c}{\textbf{Attribute}} &\multicolumn{1}{c}{\textbf{Key Relation-Attribute Chains For Reasoning}}\\
\midrule 
\multirow{2}{*}{\textbf{YG}}& latitude & (has\_capital, latitude), (located\_in, latitude), (has\_neighbor, latitude)\\
& happened\_date & (participated\_in\_inv, destroyed date), (music\_for\_inv, birth), (created\_inv, created, created date)\\
\midrule
\multirow{4}{*}{\textbf{FB}}& birth & (influenced\_by, death), (film, film\_realease), (sibling, birth), (team, team\_inv, birth)\\
& longitude & (county, longitude), (capital\_inv, longitude), (state\_province, longitude)\\
&org\_founded\_date & (member\_states, org\_founded\_date), (state, state\_province\_inv, org\_founded\_date), (nationality\_inv, death)\\
&weight & (ethnicity, ethnicity\_inv, weight), (team, team\_inv ,weight), (athlete\_inv, athlete, weight)\\

\bottomrule
\end{tabular}
\end{table*}

\begin{figure*}[ht]
    \centering    \includegraphics[width=0.90\linewidth]{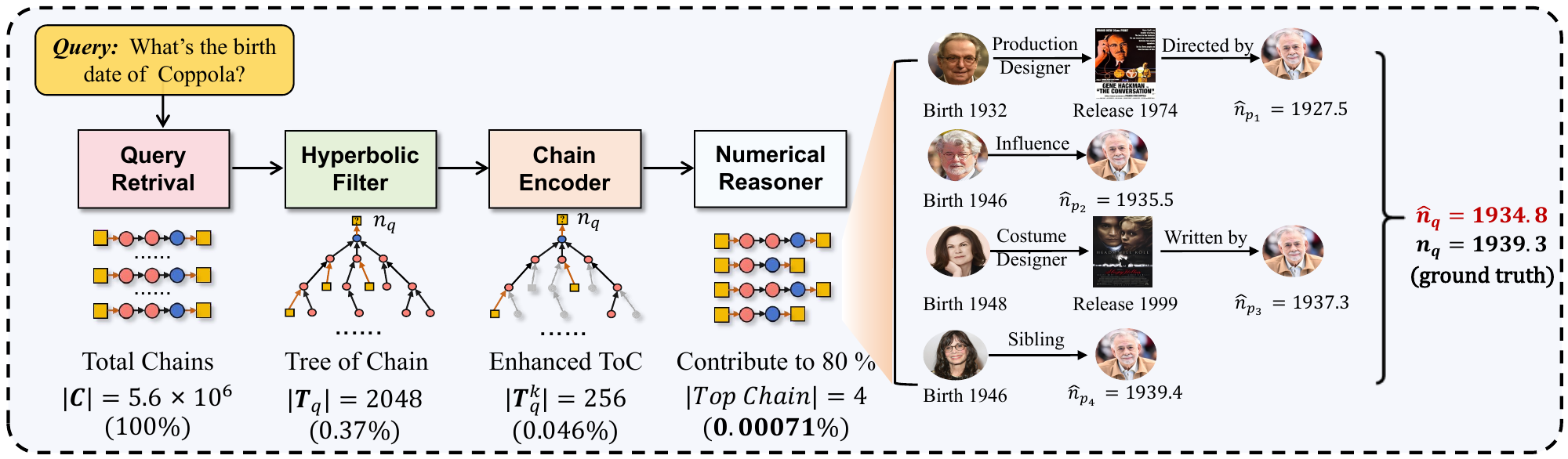}
    \vspace{-10pt}
    \caption{A case study for ChainsFormer's reasoning process on Francis Ford Coppola's birth date.}
    \label{fig:case}
    \vspace{-10pt}
\end{figure*}

To evaluate the benefits of the enhanced reasoning depth, we designed experiments across varying reasoning settings, including single-hop versus multi-hop paths and reasoning with single versus multiple attribute types. The results in Figure~\ref{fig:extensible} demonstrate that as the number of hops increases, the performance significantly improves. This is because expanding the reasoning depth from single hop to multiple hops paths enables the model to capture hidden links in incomplete knowledge graphs. Furthermore, reasoning with multiple attributes also leads to additional error reduction, highlighting the effectiveness of chains in integrating diverse information.

\subsubsection{Transparent, Step-by-Step Reasoning}

The chain-based approach precisely profile and explicitly capture reasoning patterns, enabling the identification of key RA-Chains. To assess the transparency of RA-Chains, we designed two experiments:

\textbf{Comprehensive Evaluation of Reasoning Path Transparency.}
Firstly, we identify the most important RA-chains for numerical reasoning across the two datasets. For each numerical attribute, we statistically extract the RA-Chains with the highest weights in Numerical Reasoner, treating them as key chains that play a pivotal role in numerical reasoning. Due to space limitations, we randomly select several attributes from the spatial, temporal, and quantity categories for reporting, as shown in Table \ref{table:key RA-Chain}. For each attribute type, ChainsFormer successfully identifies reasonable key RA-chains among various logic chains. For instance, in FB15K, when predicting birth dates, it selects chains involving siblings' birth dates, significant individuals' death dates, and the release dates of directed movies. For YAGO15K, when predicting longitude, it selects chains related to the longitude of capitals, locations, and neighboring areas.  These examples demonstrate ChainsFormer's ability to identify key RA-Chains, ensuring traceability of inference results and enhancing the transparency of reasoning paths.

\textbf{In-depth Case Study of Reasoning Path.} 
To further investigate the transparency of ChainsFormer, we conducted a query-level analysis using the example mentioned in the introduction—"What is the birth date of F.F. Coppola?" As shown in Figure~\ref{fig:case}, there are totally $(5.6 \times 10^6)$ chains associated with Coppola. ChainsFormer firstly retrieves a Tree of Chains (ToC) containing 2048 chains (0.37\%) from the given knowledge graph based on the query. This ToC includes numerous noisy RA-Chains. ChainsFormer then applies Hyperbolic Filter to enhance the ToC, selecting 256 relevant chains (0.046\%). After that, the filtered chains are encoded via Chain Encoder and reasoned by Numerical Reasoner, which assigns weights to the input chains to highlight their importance. This process predicts Coppola's birth date as 1934.8, close to the ground truth date 1939.3.

Notably, ChainsFormer identifies four key RA-Chains contributing over 80\% impact to the reasoning for this query. These key chains utilize information such as the birth dates of Coppola's siblings and collaborators to make a transparent and reasonable prediction. This case study highlights how ChainsFormer effectively selects the most important RA-Chains, improving numerical reasoning performance while providing consist and transparent reasoning paths.

\subsection{Effectiveness of Model Components (\textit{RQ3})}
\label{model_analysis}

\subsubsection{The Effectiveness of Hyperbolic Filter}
To analyze the effectiveness of Hyperbolic Filter, we propose the following two questions and conduct an in-depth study.

\begin{figure}[ht]
    \centering
    \includegraphics[width=\linewidth,]{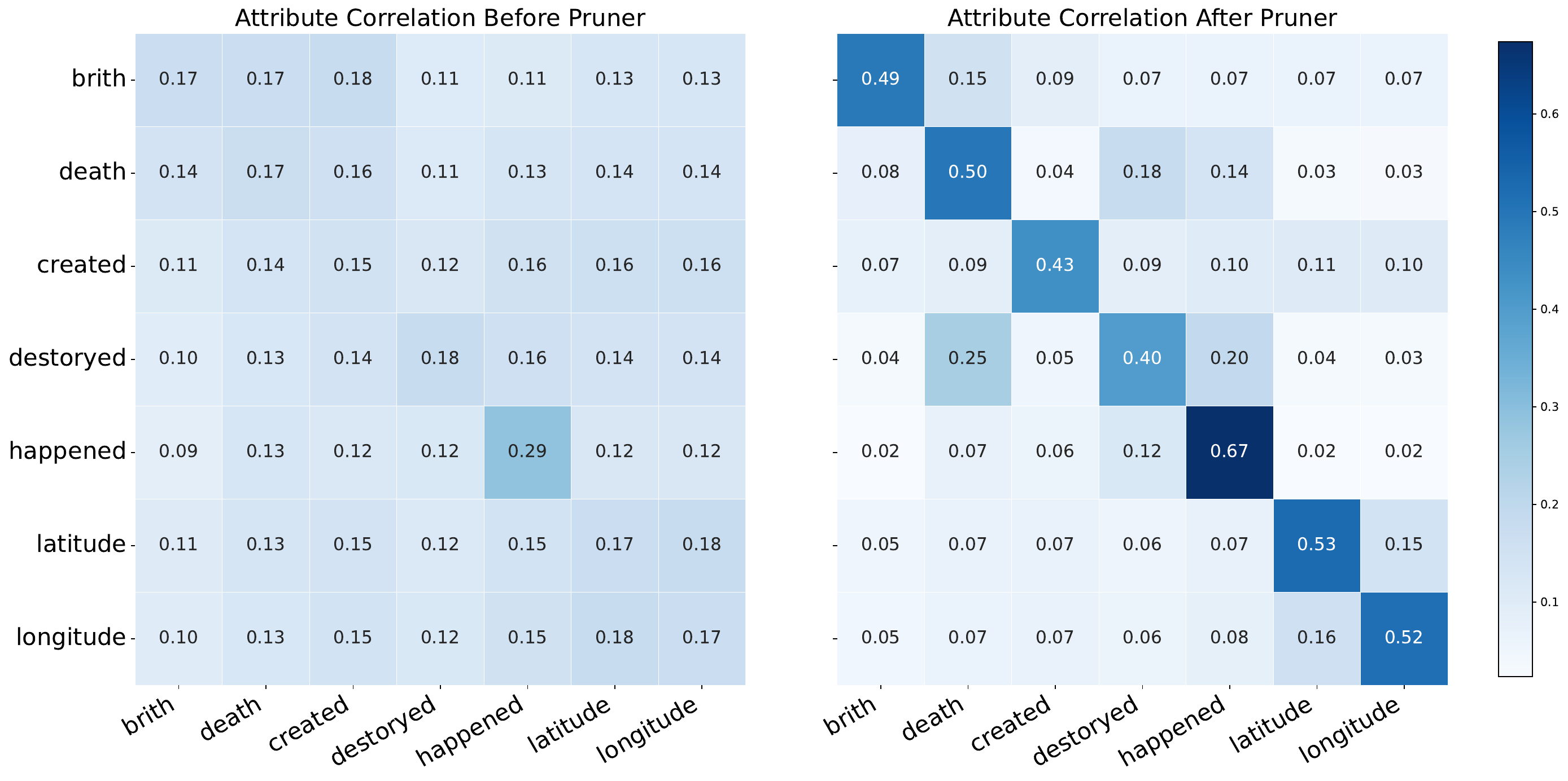}
    \vspace{-10pt}
    \caption{A case study on visualization of the effect of Hyperbolic Filter on YAGO15K dataset. Hyperbolic Filter will make the model to select the same or similar attribute for numerical reasoning.}
    \label{fig:pruner}
    \vspace{-5pt}
\end{figure}

\textbf{Whether Hyperbolic Filter can select relevant RA-Chains?} We analyzed the proportion of relevant attributes in ToC before and after filter. As depicted in Figure~\ref{fig:pruner}, the Hyperbolic Filter mainly selects the exactly same attributes with those in query. Additionally, it selects adjacent relevant attributes, such as latitude and longitude. For temporal attributes, the filter also demonstrates a preference for selecting chains within temporal domain, further confirming its effectiveness.

\begin{figure}[ht]
    \centering
    \includegraphics[width=0.9\linewidth,]{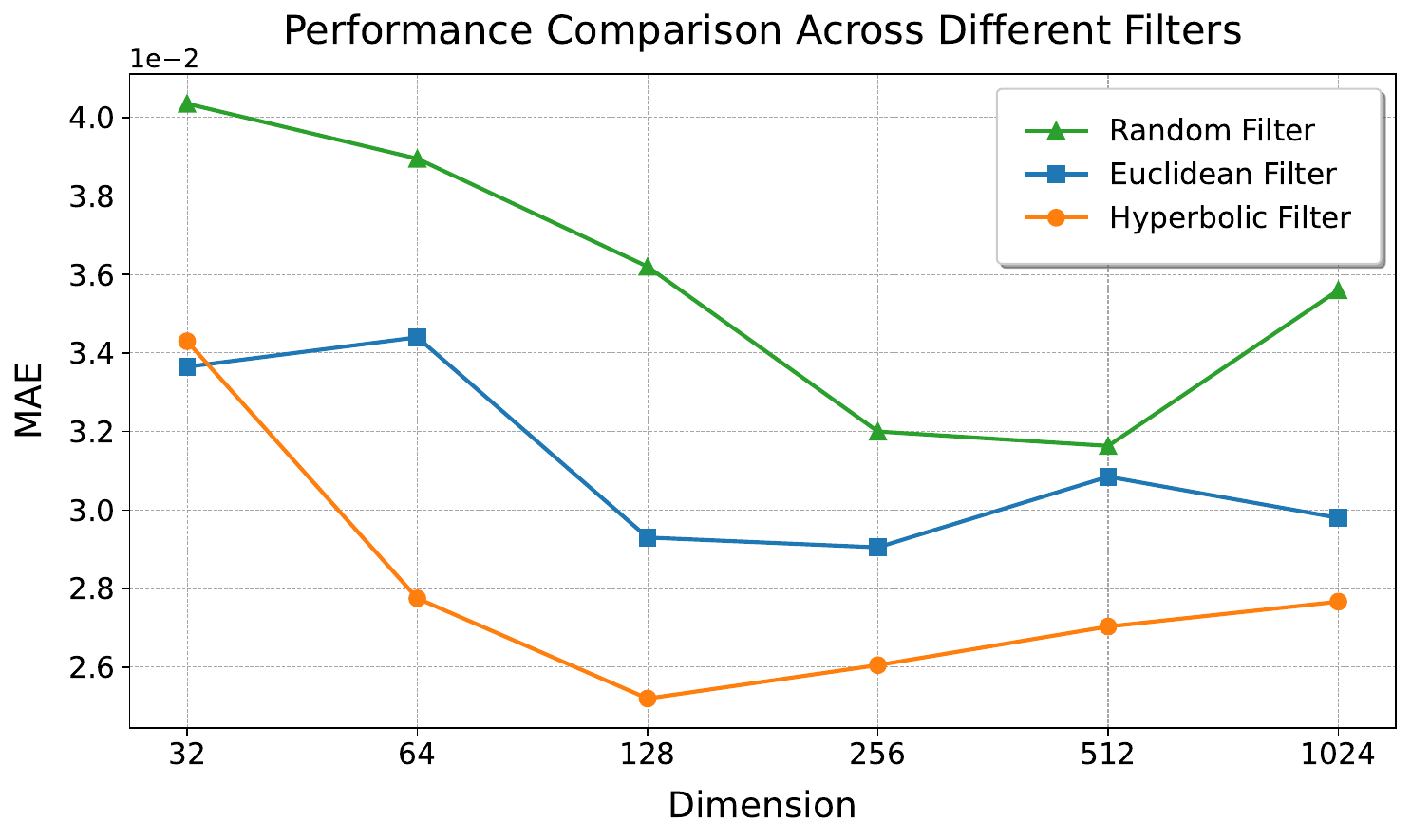}
    \vspace{-10pt}
    \caption{Performance comparison on FB15K Dataset of filtering across dimensions in hyperbolic space, Euclidean space, and random sampling.}
    \label{fig:filter_comparison}
    \vspace{-10pt}
\end{figure}

\textbf{Is filtering in hyperbolic space more effective?}
To further analyze the effectiveness of filtering in hyperbolic space,  we compared the performance of filtering in hyperbolic space, Euclidean space, and random selection across dimensions ranging from low (32) to high (1024). As illustrated in Figure~\ref{fig:filter_comparison}, hyperbolic space consistently outperforms the other approaches. Notably, even at a lower dimension (e.g., 64), hyperbolic filtering surpasses the performance of Euclidean space filtering at higher dimensions. This demonstrates the capability of hyperbolic space to effectively represent the hierarchical structure of ToC, enabling more efficient and accurate filtering.

\subsubsection{Ablation Study}

To verify the effectiveness of different modules, we conduct several degenerate variants for ablation studies. Due to space limitations, we report the normalized average performance in Table~\ref{tab:ablation}.
\begin{itemize} 
    \item \textbf{w/o Hyperbolic Filter}: Removes the Hyperbolic Filter, randomly sampling reasoning paths. 
    \item \textbf{w/o Chain Encoder}: Removes the Chain Encoder and represents RA-Chains using the average embedding. 
    \item \textbf{w LSTM as Chain Encoder}: Replaces the Transformer-based Chain Encoder with an LSTM-based model. 
    \item \textbf{w/o Numerical-Aware}: Removes the Numerical-Aware Affine Transfer module. 
    \item \textbf{w Numerical-Aware by Log}: Modifies the mapping in Numerical-Aware Affine Transfer from Float to Log. 
    \item \textbf{w/o Numerical Projection}: Removes the numerical projection in Numerical Reasoner, directly predicting based on chain embeddings. 
    \item \textbf{w/o Chain Weighting}: Removes the Chain Weighting mechanism in the Numerical Reasoner, evaluating chain importance and reliability independently. 
    \end{itemize}
    
\begin{table}[ht]
\centering
\caption{Different ablation variant's result.}
\label{tab:ablation}
\begin{tabular}{lccccc}
\toprule
\multirow{2}{*}{\textbf{Ablation Variants}} & \multicolumn{2}{c}{\textbf{YAGO15K}}& \multicolumn{2}{c}{\textbf{FB15K-237}}\\
\cmidrule(lr){2-3} \cmidrule(lr){4-5}\
& MAE & RMSE & MAE & RMSE\\
\midrule

w/o Hyperbolic Filter & 0.0185 & 0.0503 & 0.0316 & 0.0650\\
w/o Chain Encoder & 0.0213 & 0.0577 & 0.0423 & 0.0814\\
w LSTM as Chain Encoder &0.0181 & 0.0496 & 0.0264 & 0.0687 \\
w/o Numerical-Aware & 0.0167 & 0.0484 & 0.0279 & 0.0642\\
w Numerical-Aware by Log & 0.0170 & 0.0499 & 0.0295 & 0.0684\\
w/o Numerical Projection & 0.0206 & 0.0538 & 0.0421 & 0.0841\\
w/o Chain Weighting & 0.0175 & 0.0471 & 0.0276 & 0.0629\\
\midrule
ChainsFormer(Ours)& \textbf{0.0160} & \textbf{0.0467} & \textbf{0.0252} & \textbf{0.0583}\\
\bottomrule
\end{tabular}
\end{table}

As shown in Table \ref{tab:ablation}, any variant exhibits performance degradation, demonstrating the overall effectiveness of ChainsFormer. Specifically, the inclusion of the Chain Encoder significantly enhances performance, highlighting the importance of encoding RA-Chains. Moreover, directly predicting attributes using hidden features leads to severe performance drops, emphasizing the critical role of selecting an appropriate numerical projection method. Additionally, the Hyperbolic Filter's augmentation of ToC data proves highly beneficial for numerical reasoning, further showcasing its effectiveness.

\subsection{Hyperparameter Analysis and Explore Experiments \textit{(RQ4)}}

We further investigate the impact of different hyperparameters and numerical projection methods on numerical reasoning. We also conduct simple exploratory experiments with Large Language Models for comparison.
\begin{figure*}
    \centering
    \includegraphics[width=\linewidth]{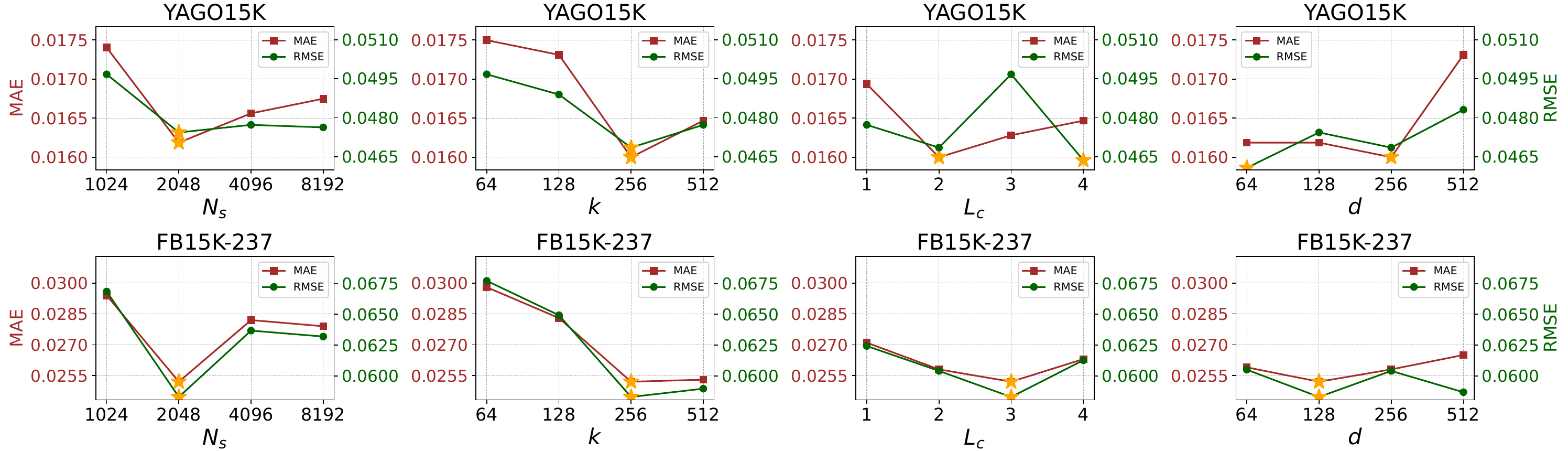}
    \vspace{-0.5cm}
    \caption{Hyperparameter study on YAGO15K and FB15K-237, exploring effects of retrieval numbers $N_s$, filtering chain numbers $k$, Transformer layers $L_c$, and hidden dimensions $d$. Brown and green represent MAE and RMSE, respectively.}
    \label{fig:hyperparameter}
    \vspace{-0.5cm}
\end{figure*} 

\subsubsection{Hyperparameter Analysis}
We investigate the hyperparameter sensitivity of ChainsFormer, including retrieval numbers $N_s$, Hyperbolic Filter's select parameter k, hidden dimension $d$ and layers of Chain encoder $L_c$.

First, we explore the impact of the number of retrieval chains used to generate the Tree of Chain, testing with \{1024, 2048, 4096, 8192\} total chains. Figure~\ref{fig:hyperparameter} shows that the number of CoT does not significantly affect the overall numerical reasoning performance. Based on our experiments, we set this parameter to 2048.

Next, we analyze the number of chains selected by the hyperbolic filter, testing with \{64, 128, 256, 512\} RA-Chains. As shown in Figure \ref{fig:hyperparameter}, selecting 256 chains achieves optimal performance. Reducing the number of selected chains significantly degrades performance, while increasing the number of paths beyond this does not enhance it, demonstrating the efficacy of the Hyperbolic Affinity Score.

Additionally, we observe the effects of the number of Transformer layers $L_c$ and the hidden dimension $d$. Two to three layers of Transformers in the Chain Encoder and Numerical Reasoner achieves the best performance across different datasets, with no further performance improvement from additional layers. Similarly, the model shows low sensitivity to the hidden dimension parameter. Based on our experiments, we set it to \{256, 128\}.

\subsection{Numerical Projection Methods}

To further investigate numerical reasoning tasks, we focus on analyzing the impact of various numerical projections in Tree Aggregator. We test several simple Numerical Projection Methods and hope more approaches can be proposed to tackle numerical reasoning challenges. For different numerical prediction methods, as shown in Table \ref{tab:operation}, the scaling projection yields the best performance. For scaling can accommodate values of any magnitude, making it more versatile. In contrast, direct regression based on embedding produces the poorest results, particularly obvious in FB15K-237, which contains more complex attribute values.

\begin{table}[ht]
\centering
\caption{Different numerical prediction projection methods.}
\label{tab:operation}
\begin{tabular}{lccccc}
\toprule
\multirow{2}{*}{\textbf{Projection}} & \multicolumn{2}{c}{\textbf{YAGO15K}}& \multicolumn{2}{c}{\textbf{FB15K-237}}\\
\cmidrule(lr){2-3} \cmidrule(lr){4-5}\
& MAE & RMSE & MAE & RMSE\\
\midrule
Direct & 0.0206 & 0.0538 & 0.0421 & 0.0841\\
Translation & 0.0173 & 0.0500 & 0.0264 & 0.0593\\
Scaling & \textbf{0.0160} & \textbf{0.0467} & \textbf{0.0252} & \textbf{0.0583}\\
Combined & 0.0170 & 0.0499 & 0.0263 & 0.0584\\

\bottomrule
\end{tabular}
\vspace{-3pt}
\end{table}

\subsection{Comparison with LLM}

With the development of large language models (LLMs), their reasoning capabilities across various domains have shown significant improvement. To evaluate our model's performance against LLMs, we conduct a comparative experiment, as illustrated in Table~\ref{tab:LLM}. In this zero-shot setting, we provide only the RA chains and target entity attribute values as input to the LLMs. Entity semantics are removed to avoid potential misguidance from unfamiliar entity literal information~\cite{ToG} and to prevent label leakage, such as querying the latitude of the United States directly based on pre-trained data. The results show that ChainsFormer outperforms the LLMs, achieving an average improvement of 33.0\% and 26.1\% in MAE across the two datasets. This highlights Chainsformer's superior numerical reasoning capabilities

\begin{table}[ht]
\centering
\caption{Comparsion experiment with LLMs.}
\label{tab:LLM}
\begin{tabular}{lccccc}
\toprule
\multirow{2}{*}{\textbf{Models}} & \multicolumn{2}{c}{\textbf{YAGO15K}}& \multicolumn{2}{c}{\textbf{FB15K-237}}\\
\cmidrule(lr){2-3} \cmidrule(lr){4-5}\
& MAE & RMSE & MAE & RMSE\\
\midrule
ChatGPT-3.5-turbo& 0.0571 & 0.1126 & 0.0496 & 0.1123\\
ChatGPT-4.0-turbo& 0.0239 & 0.0556 & 0.0341 & 0.0957\\
\midrule
ChainsFormer(Ours)& \textbf{0.0160} & \textbf{0.0467} & \textbf{0.0252} & \textbf{0.0583}\\
\bottomrule
\end{tabular}
\end{table}

%% file: 6_conclusion.tex
\section{CONCLUSION}
\label{6_conclusion}

In this work we introduce ChainsFormer, a novel framework, to enhance numerical reasoning from a chain perspective. By constructing Relation-Attribute Chains (RA-Chain), the model effectively profiles and captures reasoning patterns, enabling a step-by-step and explicit reasoning process. Experimental results show that ChainsFormer significantly improves numerical reasoning performance, enhancing reasoning depth to multiple hops, uncovering hidden connections in incomplete knowledge graphs and providing accurate inference results.

In the future, we plan to extend ChainsFormer by integrating multimodal information, such as text and images, to further enhance its reasoning capabilities. Additionally, we will introduce a chain quality evaluation mechanism to address low-quality RA-Chains. We also aim to explore its application in broader reasoning tasks and scale the framework to larger knowledge graphs. Furthermore, we will investigate the compatibility of RA-Chains with large language models to unlock new potential in knowledge-driven reasoning.

%% file: 7_Acknowledgement.tex
\section{Acknowledgments}
\label{7_Acknowledgments}

This work was partially supported by National Key R\&D Program of
China (No.2022YFB3904204), NSF China (No.62272301, 623B2071, T2421002, 62061146002, 62020106005) and Shanghai Pilot
Program for Basic Research - Shanghai Jiao Tong University.